\documentclass[11pt]{article}

\usepackage{acl}

\usepackage{times}
\usepackage{latexsym}

\usepackage[T1]{fontenc}

\usepackage[utf8]{inputenc}

\usepackage{microtype}

\usepackage{inconsolata}

\usepackage{graphicx}
\usepackage{amssymb}
\usepackage{booktabs}
\usepackage{tabularray}

\title{Efficient One-to-Many Translation with Joint Multi-Stream Diffusion}

\author{Yiwen Guan, Jacob Whitehill \\
  Worcester Polytechnic Institute \\
  \texttt{yguan2@wpi.edu, jrwhitehill@wpi.edu} \\}

\begin{document}
\maketitle
\begin{abstract}
One-to-many machine translation (MT) is computationally expensive for autoregressive (AR) systems, which suffer from linear latency scaling with both sequence length and the number of target languages. 
We explore how diffusion can enable multilingual translation with a discrete diffusion framework that refines all target languages in parallel, achieving sublinear latency scaling with the number of targets, and supports deployment as a single unified model to replace multiple independent systems.
Conditioned on a continuous semantic anchor rather than source tokens, our framework supports zero-shot transfer to unseen source languages without retraining, maintaining approximately $75\%$ of its supervised translation quality on zero-shot sources.
We investigate the quality-latency frontier and find that with accelerated sampling, it achieves comparable supervised quality to AR baselines with a $2 \times$ speedup and $11.9\%$ better zero-shot BLEU. 
These results highlight the potential of joint multi-stream diffusion as a practical and flexible alternative for efficient one-to-many translation. 

\end{abstract}


\section{Introduction}

Real-world multilingual translation, such as in live interpreting for international meetings (e.g., the United Nations, the European Parliament), frequently requires different target languages simultaneously from a single source stream.
In these scenarios, the set of requested languages can change dynamically as different client nodes join or leave the session, demanding a flexible and low-latency framework. 
However, existing approaches predominantly rely on Transformer-based \citep{vaswani2017attention} autoregressive (AR) architectures, typically either deploying separate one-to-one models for each language pair, or exploiting a unified model managing multilingualism via language-specific control tokens \citep{johnson2017google, fan2021beyond}. 
While they attain high accuracy, AR models generate tokens sequentially and require separate decoding passes for each target language. This computational bottleneck makes AR models expensive for parallel one-to-many translation, especially in resource-constrained applications, as inference latency scales linearly with both the sequence length and the number of targets \citep{gu2018non, kasai2020deep}. 

To mitigate this issue, non-autoregressive (NAR) translation has been recently investigated \citep{gu2018non, ghazvininejad2019mask, xiao2023survey}.  
By generating tokens in parallel, NAR methods offer compelling speed-quality trade-offs \cite{qian2021glancing}.
However, most existing methods assume conditional independence among target tokens, resulting in the ``multimodality problem'' \citep{gu2018non}. 
Furthermore, naively adapting these frameworks to one-to-many translation (e.g., running multiple models in parallel) is both memory intensive and computationally inefficient. 

As an alternative, diffusion models (DMs) have emerged as a powerful NAR paradigm for sequence generation through iterative parallel refinement \citep{ho2020denoising}. 
Continuous DMs map tokens into a continuous embedding space \citep{li2022diffusion, gong2023diffuseq}, while discrete DMs operate directly on the categorical vocabulary \citep{austin2021structured, nie2025large}. 
Despite their potential, applying DMs to machine translation (MT) remains underexplored. 
XDLM \cite{chen2023xdlm} uses discrete diffusion with cross-lingual pretraining but is limited to one-to-one tasks. 
Scaling such diffusion frameworks raises important design questions of the optimal input representation, attention mechanism, and diffusion inference schedule.


In this work, we explore the viability and design space of parallel multilingual MT. As an instantiation, we propose \textbf{PrismDiff}, a discrete diffusion framework for parallel one-to-many generation. Metaphorically, PrismDiff acts like a prism: it ``refracts'' a shared source representation into multiple target outputs through parallel diffusion refinement. 
In multilingual MT tasks, PrismDiff harnesses a language-agnostic semantic anchor to guide the diffusion process, rather than conditioning directly on source tokens. 
This shared anchor not only enables zero-shot transfer capability to unseen source languages but also improves translation accuracy.

Our contributions are summarized as follows: 
\begin{enumerate}
\setlength\itemsep{0em}
    \item We propose PrismDiff, a multi-stream discrete diffusion framework for one-to-many translation, which refines multiple target languages in parallel from a shared semantic anchor. 
    Joint optimization over all target streams provides implicit cross-lingual regularization, achieving better translation quality than running independent one-to-one MT systems. 
    \item We explore the quality-latency frontier of diffusion systems against AR baselines. We show that under supervised settings, PrismDiff achieves a competitive frontier and outperforms AR systems at matched low-latency settings. Furthermore, under source-side zero-shot settings, the framework forms a stronger performance frontier than AR baselines. 
    
    \item 
    Through systematic ablations, we show that effective zero-shot transfer requires a high-quality cross-lingual semantic space rather than an arbitrary multilingual encoder, and the anchor mechanism is not replaceable by increasing the number of diffusion steps alone. 
    
\end{enumerate}


\section{Related Work}
\subsection{Non-autoregressive Translation}
Non-autoregressive translation (NAT) aims to reduce the sequential bottleneck of autoregressive decoding by predicting target tokens in parallel \citep{gu2018non,guo2019non,ghazvininejad2019mask}. 
This set of methods has produced strong speed-quality trade-offs, especially through conditional masked language modeling and iterative refinement. 
Several strong NAT systems have been proposed for one-to-one translation, for example, conditional masked language models repeatedly update masked positions conditioned on the source sentence \citep{ghazvininejad2019mask}, while glancing training exposes the model to a curriculum of partially observed target tokens \citep{qian2021glancing}. 
Although these methods establish compelling quality-latency trade-offs in bilingual settings, they are not directly designed for parallel one-to-many generation, and do not provide a natural mechanism for sharing a single semantic representation across multiple target streams. Similarly, XDLM \cite{chen2023xdlm} explores discrete diffusion for cross-lingual generation  focusing on one-to-one translation. Our work targets different settings: a unified model decoding into multiple target languages in parallel, which motivates different design choices and baselines. 
Previous systems for multilingual translation rely on task prompts or specialized architectures to coordinate target languages \citep{johnson2017google,fan2021beyond,azpiazu2020framework,guan2025transformer}, which are limited in deployment flexibility. 
In contrast, our work uses a single diffusion canvas in which multiple target streams are optimized together, allowing for joint one-to-many decoding. 

\subsection{Diffusion Language Model}
In the context of language modeling, diffusion language models (DLMs) generally fall into two categories: continuous and discrete. 
Continuous DLMs operate in embedding space and learn to denoise latent token representations \citep{li2022diffusion,gong2023diffuseq}. 
Discrete DLMs instead define the corruption and reverse processes directly over categorical tokens, often using masking or transition matrices \citep{austin2021structured,he2023diffusionbert,nie2025large,bie2025llada2,ye2025dream}. 
Our method follows the discrete masked diffusion paradigm, which naturally supports parallel token reconstruction and variable sampling budgets.

Diffusion-based translation remains less explored than AR and other NAT methods. 
XDLM \citep{chen2023xdlm} studies cross-lingual discrete diffusion with pretraining, but focuses primarily on one-to-one generation. 
In contrast, we target parallel one-to-many translation, where the model must generate multiple target languages for the same source sentence concurrently; we also emphasize the quality-latency frontier exposed 
by changing the number of sampling steps on the same model. 

\subsection{Inference-time Acceleration for Diffusion Models}
A practical limitation of DMs is that generation quality depends on the number of denoising steps, creating a trade-off between latency and output quality. 
Prior work has explored techniques to accelerate discrete diffusion inference speed while maintaining quality \citep{chen2025dpad,liu2025dllm,wu2025fast, bartosh2026forward}, including training-free acceleration strategies, such as non-uniform schedules and jumpy sampling \citep{yeh2024cross,chen2024fast,irwin2025semlaflow}. 

In our work, we adopt the logarithm-uniform schedule applied in \citealt{irwin2025semlaflow} as a control knob at inference time, allowing a single trained model to operate at multiple quality-latency points by varying the number of sampling steps, without modifying model capacity or retraining the model. 
This is the key mechanism we use to compare with AR baselines, where latency reduction requires shrinking model depth.


\begin{figure*}[ht]
\centering
  \includegraphics[width=1.0\textwidth]{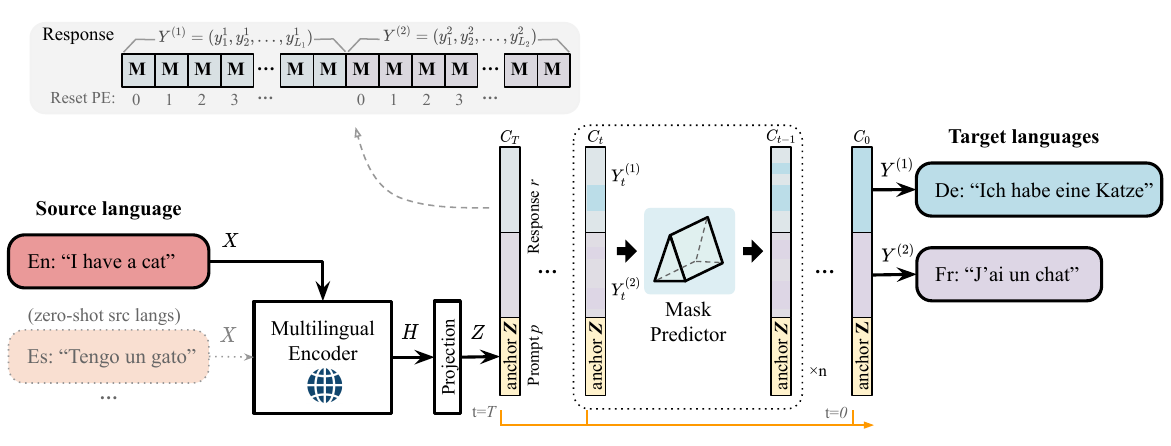}
\vspace{-7mm}
  \caption{\textbf{An illustration of PrismDiff.} ``Reset PE'' is short for Reset Positional Encoding. PrismDiff takes one source language sentence $X$ as input, generating multiple target languages $\{Y^{(1)},^{(2)},\dots,Y^{(K)}\}$ concurrently. The source language can be replaced with other zero-shot languages, such as Spanish. 
  For simplicity, we use En-\{De, Fr\} translation as an example. 
  In practice, the total diffusion refinement steps can be reduced to accelerate.}
  \label{fig:PrismDiff}
\vspace{-3mm}
\end{figure*}

\section{Methodology}

\subsection{Preliminaries} 
We adopt the framework of existing discrete diffusion models, which generate outputs through iterative denoising of masked sequences \citep{austin2021structured, nie2025large, bie2025llada2}. 
However, we focus on one-to-many generation, conditioned on a sequence of latent representations instead of source tokens. 

Given a source sequence $X$ with length $L_\mathrm{src}$, our goal is to generate $K$ target translations $\mathcal{Y}=\{Y^{(1)}, Y^{(2)},\dots,Y^{(K)}\}$ in parallel, where the $k$-th target language sequence $Y^{(k)}=(y_1^{k},\dots,y_{L_k}^{k})$ consists of discrete tokens from a vocabulary $\mathcal{V}$. 
Following existing discrete DLMs, we represent each example as a concatenate a prompt $p$ (conditioned information) and a response $r$ (target to be generated). In our setting, $p$ remains unmasked, while $r$ is progressively corrupted and reconstructed.

The forward process is a discrete masked corruption process defined over $T$ discrete timesteps $t\in \{ 1,\ldots,T\}$, where $T$ controls the total amount of noise applied during training. 
Let $r_0$ be the clean response and $r_t$ be the corrupted response at timestep $t$, where 
 $r_{t,i}$ denotes the token at the position index $i$ in the generated sequence at diffusion timestep $t$. 
The forward process independently applies $\texttt{[MASK]}$s to the clean response $r_0$ to produce the corrupted state $r_t$, based on a schedule $\gamma(t)$ (i.e., the probability of each token being masked at timestep $t$), where $\gamma(T)=1$ results in fully masked sequence. 

The reverse process employs a neural network $p_\theta(\cdot)$ to denoise the sequence, which takes the concatenation $[p;\ r_t]$ of the prompt $p$ and the corrupted sequence $r_t$, and predicts the probability of the original response token $p_\theta(r_{0,i}\ |\ p,\ r_t)$ for each masked position $i \in \mathcal{M}_t$, where $\mathcal{M}_t$ denotes all masked positions at time $t$. 
The training objective is to maximize the variational lower bound of data log-likelihood, optimized via a cross-entropy loss restricted to masked tokens: 
\begin{equation}
\label{eq:loss}
    \mathcal{L}(\theta) \triangleq -\mathbb{E}_{t,p,r_0,r_t} 
    \left[ 
    \sum_{i\in \mathcal{M}_t} \, \mathrm{log} \, p_\theta \big(r_{0,i}\ |\ p,\ r_t \big) 
    \right]
\end{equation}

The inference stage begins from a fully masked sequence, and iteratively denoises it over multiple refinement steps. 
At each step, the model first predicts the clean sequence from the current noisy state, then performs a remasking process that selects a fraction of tokens to be masked again, ensuring the sequence becomes progressively less noisy.

\subsection{Proposed Method}
\label{sec:proposed_method}
An illustration of our framework is in Figure~\ref{fig:PrismDiff}.  

\paragraph{Language-Agnostic Semantic Anchor.}
Conditioning on source tokens would bind the model to a specific source language, preventing zero-shot transfer. 
To decouple the generation process from source language surface forms and enable zero-shot transfer, 
we condition the diffusion process on a continuous representation rather than on source tokens directly. 
Specifically, we extract the last hidden states $H\in\mathbb{R}^{L_\mathrm{src}\times d_\mathrm{model}}$ from the source input $X$ using LaBSE (Language-agnostic BERT Sentence Embedding) \citep{feng2022language}, a pretrained multilingual encoder optimized to align parallel sentences across languages into a shared embedding space. 
Then $H$ is projected into the diffusion model's dimension $d_\mathrm{model}$ via a learnable linear projection to obtain the anchor $Z\in \mathbb{R}^{L_\mathrm{src}\times d_\mathrm{model}}$.

\paragraph{Multi-Stream Generation.} 
To enable parallel generation, we construct a unified ``canvas'' $C_t$ at timestep $t\in[0, T]$. This canvas concatenates the anchor $Z$ with all $K$ noisy target streams:
$C_t=[p_0;\ r_t]=[Z;\ Y_t^{(1)};\ \dots;\ Y_t^{(K)}]$  
where $Y_t^{(k)}$ represents the $k$-th target language stream at diffusion step $t$. 
The neural network processes the entire $C_t$ simultaneously and outputs predictions for all masked positions across all streams. 
Masked tokens in the $k$-th stream are replaced by a language-specific mask token $\texttt{[MASK]}_k$, added as special tokens to the shared BPE vocabulary, rather than a generic mask. 

Full cross-stream attention introduces interference between target languages during generation. 
We therefore replace full cross-stream attention with exclusive attention within each stream. 
While the attention is isolated, 
all streams share the same model parameters and are jointly optimized, providing implicit cross-lingual regularization that benefits each individual stream. 

Since the canvas concatenates multiple sequences, standard positional encoding would assign monotonically increasing indices. Instead, we employ Reset Positional Encoding ($E_\mathrm{pos}$), as illustrated in the upper left corner of Figure~\ref{fig:PrismDiff}. 
For the anchor $Z$, position indices range from 0 to $L_{src}-1$. 
For each target stream $Y^{(k)}$, we restart the position indices from 0 to $L_{tgt}-1$. 
This design allows for the dynamic reordering, removal or addition of target languages without affecting performance, and encourages the model to correctly identify token positions within individual streams. 

\paragraph{Training.} 
During training, we sample a timestep $t\sim \mathcal{U}(1,T)$ and corrupt the target streams, while the anchor $Z$ remains uncorrupted. 
The model predicts the original tokens from all masked positions across all streams in parallel. 
The loss function is adapted from Eq. \ref{eq:loss}: 
\begin{equation}
\label{eq:loss_train}
    \mathcal{L}(\theta) = -\mathbb{E}_{t,Z,r_0,r_t} 
    \left[ 
    \sum_{i\in \mathcal{M}_t} \, \mathrm{log} \, p_\theta \big(r_{0,i} | Z, r_t \big) 
    \right]
\end{equation}
where $r_t=[Y_t^{(1)};\ \dots;\ Y_t^{(K)}]$ is the concatenation of all target streams. 

\paragraph{Inference.} 
In the inference phase, we start with a fully masked canvas $C_T$ with length $L$ containing only the anchor $Z$ and language-specific masks. 
At each refinement step, the model retains the top $(1-m)\times L$ most confident tokens and remasks the remainder, where the masking ratio $m$ is determined by the schedule. 
We experiment on both the vanilla schedule and a logarithm-uniform schedule following \citealt{irwin2025semlaflow} to progressively unmask tokens. 
Specifically, the logarithm-uniform mask distribution over the masking ratio $m$ is defined as: 
\begin{equation}
\label{eq:log_schedule}
  \log m \sim \mathcal{U}\left(\log \frac{1}{L}, 0\right)
\end{equation}
, the masking ratio $m$ is bounded within $[1/L, 1]$, and $1/L$ corresponds to the single-token mask limit. 
See more details in \ref{appendix:log_schedule}.

Since discrete diffusion models lack explicit autoregressive constraints, they may generate consecutive repetitive tokens.
To mitigate this, we apply Connectionist Temporal Classification (CTC) decoding \cite{graves2006connectionist} as a post-processing step without training on CTC loss. Specifically, we use the CTC collapse rule (removing consecutive duplicates and blanks) on the final generated sequence, which effectively acts as a structural regularization for NAR output.


\section{Experimental Setup}

\subsection{Baselines}

We compare PrismDiff with the following AR and diffusion baselines: 

\begin{itemize}
\setlength\itemsep{0em}
    \item \textbf{AR}: Decoder-only AR Transformers conditioned on the same source-side representations (anchor $Z$), and trained on all En-X pairs with task prompts. 
    We vary the number of decoder layers to trace the AR quality-latency frontier. 
    Our preliminary experiments (detailed in \ref{appendix:results}) show that this architecture outperforms other AR settings such as encoder-decoder. 
    \item \textbf{Diff-Indep}: Independent diffusion decoders conditioned on the same anchor $Z$, but trained separately for each En-X pair. This isolates the contribution of joint multi-stream optimization from the anchor mechanism. 
    \item \textbf{Diff-SrcTokens}: A multi-stream diffusion model identical to PrismDiff, but conditioned on source tokens instead of anchor $Z$. This tests whether the gains of PrismDiff come from anchor-based semantic conditioning. 
    \item \textbf{Transformer-Encoder Trees (TET)} : A strong concurrent one-to-many NAT model \cite{guan2025transformer} that generates multiple targets with a unified encoder-tree architecture with CTC. TET is trained on En-X tasks but lacks zero-shot transfer capability. 
\end{itemize}

We do not directly compare against one-to-one NAT systems such as GLAT and XDLM in a one-to-many setting, as our focus is on isolating specific design decisions within a unified diffusion framework.
The Diff-Indep is designed for controllable comparisons: it uses the same architecture and anchor as PrismDiff but generates each languages independently.

\subsection{Datasets and Evaluation Metrics}
\paragraph{Datasets.} We utilize Multi30K \cite{elliott2016multi30k} as our primary benchmark. Multi30K is a multilingual dataset of image descriptions with short, visually grounded sentences. We use the 
English-to-German and English-to-French
(En-\{De, Fr\}) pairs (29k) for training, and 1k for testing. To evaluate zero-shot transfer performance, we create a synthetic Spanish-to-X (Es-X) test set by translating English into Spanish using NLLB-200 \citep{costa2022no}. 

To further verify the effectiveness of anchor mechanism, we conduct supplementary experiments on Europarl \citep{koehn2005europarl}, a large-scale, aligned parallel corpus from European Parliament proceedings. We construct a training set of 260k 
English-to-French, Dutch, Romanian, and Danish examples (En-\{Fr, Nl, Ro, Da\}), and 33k for testing, with Spanish serving as the unseen source language. 
Europarl's Spanish evaluation set comes from original human-authored corpus, allowing us to rigorously verify our zero-shot transfer claims. 

\paragraph{Metrics.} Translation quality and cross-lingual semantic transfer are measured with SacreBLEU\footnote{The SacreBLEU signature is \path{nrefs:1|case:mixed|eff:no|tok:13a|smooth:exp|version:2.5.1}.} \citep{post2018call} and COMET\footnote{COMET is with \url{Unbabel/wmt22-comet-da} (v2.2.7)} \citep{rei2020comet}. Inference efficiency is reported as the average end-to-end wall-clock latency (ms) for generating all requested target sentences for a source sentence on a single model instance. 
Note that AR systems can reduce wall-clock latency by running multiple model replicas in parallel, which increases memory and deployment complexity. Our comparison focuses on one-to-many latency in a single model.

\subsection{Implementation Details}

All diffusion systems use an 8-layer Transformer encoder backbone ($d_{model}=768$, $n_{head}=8$) to serve as the diffusion decoder; each contains about 230M parameters. 
Unless otherwise noted, the diffusion systems use a total number of diffusion steps $T=100$ on Multi30K and $T=40$ on Europarl with accelerated sampling at different numbers of sampling steps. This allows a single trained model to exhibit multiple operating points by adjusting the number of sampling steps $t$. We also report $T=10$ diffusion results as supporting comparisons. We set the max length of diffusion blocks  $L_\mathrm{src}=40$ for Multi30K and $L_\mathrm{src}=80$ for Europarl. 

The decoder-only AR models use similar Transformer settings, contain about 247M, 216M, and 208M parameters in the 8-layer, 4-layer, 3-layer versions, respectively. AR models use the same max length as diffusion systems. 

The Transformer-Encoder Tree (TET) model \cite{guan2025transformer} for Europarl has 4 leaf nodes (corresponding to the targets), each with a depth of 6. It contains 16 distinct Transformer encoder layers in total, distributed across its tree topology,  resulting in a model size of 556M. 

LaBSE is kept frozen during training; only the linear projection layer is trained as an adapter. The models are optimized using AdamW \citep{loshchilovdecoupled} with a learning rate of 5e-5. Models are trained for 60 epochs, and models with the best validation loss are used for evaluation. 
All models are trained and evaluated on a single NVIDIA L40S GPU, using batch size 8 during training and batch size 1 for evaluation. 

\paragraph{Decoding Protocol.}
All models use a BPE vocabulary across all target languages. 
In the inference phase of all diffusion models, CTC collapse (as mentioned in Sec. \ref{sec:proposed_method}) is applied after tokenization decoding. 
We find this technique consistently outperforms (by an average of 2.5\%) vanilla decoding methods by removing consecutive duplicate tokens at the subword level. See \ref{appendix:results} for detailed results. 
For both diffusion and AR systems, we truncate each decoded sequence at the first EOS token.


\begin{table}[t]
\centering
\setlength{\tabcolsep}{3pt}
\resizebox{0.95\columnwidth}{!}{%
  \begin{tabular}{clc|cc|r}
    \toprule
     \textit{Src.}&\textbf{Model}  &$t$& \textbf{BLEU}  &\textbf{COMET}&\textbf{Latency}\\
    \midrule
  \textit{En}
&Diff-Indep&50& 35.95& 0.710&128.8 \\ 
  
&&30& 35.55& 0.710&76.6\\ 
  
&&25& 35.60& 0.705&63.5 \\  
  \cmidrule{2-6}
  
    &Diff-SrcTokens&50& 35.03& 0.695&122.5\\
  
& &30& 35.07& 0.690&72.6\\

& &25& 34.40&0.685& 60.2\\
  \cmidrule{2-6}
     
&AR (8-layer)&& \textbf{37.25}&\textbf{0.720}&148.3\\

&AR (4-layer)&& 36.03&\textbf{0.720}&92.5\\
     
&AR (3-layer)&& 35.48&0.710&76.0\\
  \cmidrule{2-6}
  
&\textbf{PrismDiff}&50 & 37.06& \textbf{0.720}&147.2 \\
  
& &30 & \textbf{37.09}& 0.715&83.2 \\

& &25& \textbf{36.97}& \textbf{0.715}&68.7 \\
    \midrule \midrule
  
\textit{Es}
&Diff-Indep &50 & 27.21& 0.670 &128.5 \\

& &30 & 26.88& 0.670 &76.5 \\

& &25& 26.94& 0.670 &63.7\\
  \cmidrule{2-6}
     
&Diff-SrcTokens &50 & 0.14&0.305&122.8\\
 
& &30 & 0.12&0.300&72.5\\
     
& &25& 0.25&0.300&62.5\\
  \cmidrule{2-6}
 
&AR (8-layer) & & 25.22&0.655&154.5\\

&AR (4-layer) & & 26.03& \textbf{0.670}&92.5\\
  
&AR (3-layer)& & 24.54& 0.655&77.6\\
  \cmidrule{2-6}
  
 &\textbf{PrismDiff}&50 & \textbf{28.23}& \textbf{0.674}&141.7 \\
  
& &30 & \textbf{28.30}& \textbf{0.670}&83.1 \\

& &25 & \textbf{28.22}& \textbf{0.670}&69.3\\
    \bottomrule
  \end{tabular}
}
\vspace{-2mm}
\caption{\textbf{Main results on Multi30K.} Metrics are averaged on De and Fr. Es is a zero-shot source language. $t$ is the accelerated sampling steps using logarithm-uniform scheduling. All diffusion systems are trained with $T=100$. Latency is reported in ms. Bold text denotes the best results under comparable latency budgets (or equivalent sampling steps $t$). See \ref{appendix:results} for details. }
\label{tab:multi30k}
\vspace{-2mm}
\end{table}

\begin{table}[t]
\centering
\setlength{\tabcolsep}{3pt}
\resizebox{0.9\columnwidth}{!}{%
  \begin{tabular}{clc|cc|r}
    \toprule
     \textit{Src.} 
&\textbf{Model}  &  $t$& \textbf{BLEU}  &\textbf{COMET}&\textbf{Latency}\\
  \midrule
     \textit{En} 
&TET&  & 26.64& 0.630& 12.1\\
  \cmidrule{2-6}
  
 & Diff-SrcTokens       & 40& 26.23& 0.728&224.8\\
  
& & 20& 24.48& 0.735&93.7\\
  \cmidrule{2-6}
  
&\textbf{PrismDiff}       & 40& 29.54& 0.778&239.3\\
     
& &   20& 28.84&0.785&112.7\\
    \midrule \midrule
    \textit{Es} 
&TET&  & 0.14& 0.265& 11.4\\
  \cmidrule{2-6}
  
&Diff-SrcTokens       &  40
& 0.03& 0.320&225.6\\
  
& & 20& 0.33& 0.320&95.8\\
  \cmidrule{2-6}
  
&\textbf{PrismDiff}       & 40
& 23.47& 0.745&237.0\\
     & &   20& 22.10& 0.745&112.1\\
    \bottomrule
  \end{tabular}
}
\vspace{-2mm}
\caption{\textbf{Main results on Europarl.} Metrics are averaged on Fr, Nl, Ro, and Da. Es is a zero-shot source language. All diffusion systems are trained with $T=40$. 
TET generates all targets in a single CTC pass without sequential or iterative steps, thus achieving low latency. 
} 
\label{tab:euro}
\vspace{-3mm}
\end{table}

\section{Results and Analysis}
\label{sec:results}

Table \ref{tab:multi30k} and Table \ref{tab:euro} summarize the performance on Multi30K and Europarl. 
Rather than comparing a single decoding configuration, we analyze the quality--latency frontier induced by diffusion sampling steps and AR decoder depth.  
Compared with strong AR baselines, PrismDiff achieves a competitive quality--latency trade-off: it reaches comparable SacreBLEU and COMET scores with substantially lower latency, while showing advantages in source-side zero-shot transfer and parallel one-to-many generation.

\subsection{Supervised Quality-Latency Frontier}
\begin{figure*}[t]
\centering
  \includegraphics[width=1.0\textwidth]{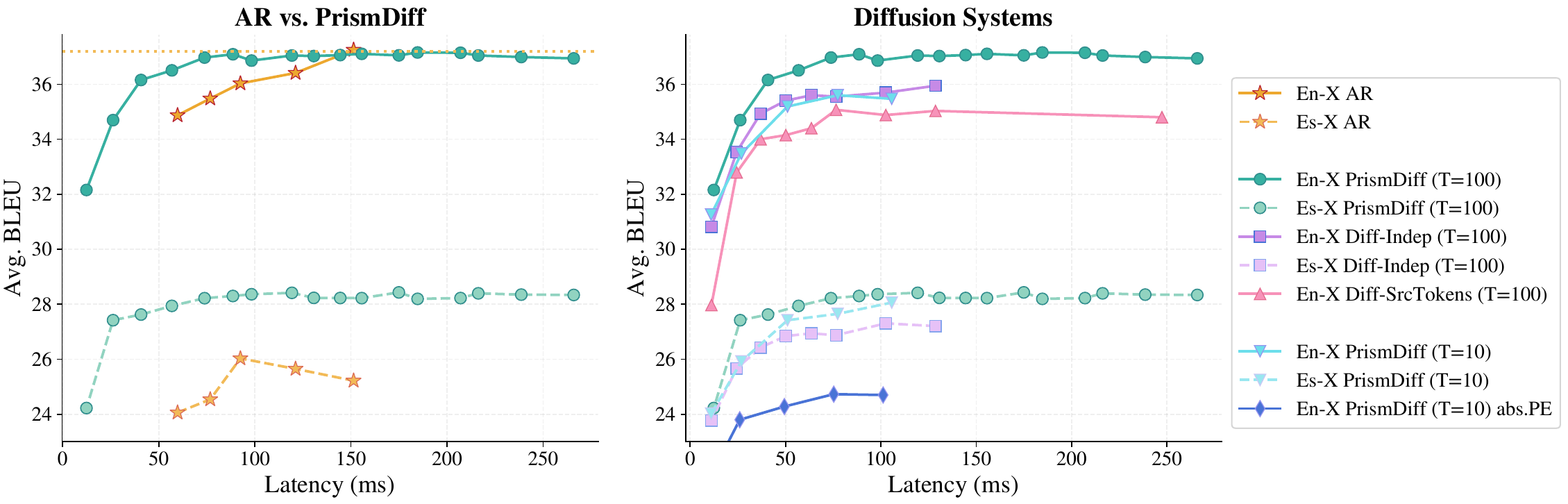}
\vspace{-6mm}
  \caption{\textbf{Quality-latency frontiers on Multi30K.} Solid curves denote supervised En-X results, and dashed curves denote source-side zero-shot Es-X results.
  \textbf{Left}: Comparison between AR models and PrismDiff. 
  \textbf{Right}: Comparison among diffusion-based systems. 
  We omit the Es-X results for Diff-SrcTokens as its BLEU is $<1$. 
  $T$ indicates the total diffusion steps, and the latency operating points are obtained by varying the number of sampling steps. 
  For AR models, operating points are obtained by varying the decoder depth from 3 to 8. 
  } 
  \label{fig:frontier}
\vspace{-3mm}
\end{figure*}

Figure~\ref{fig:frontier} (left) compares PrismDiff with strong decoder-only AR baselines. 
On supervised En-X tasks, PrismDiff reaches most of its BLEU gains within 25--30 sampling steps ($t$), corresponding to roughly 70--100 ms latency. 
Beyond this regime, additional sampling steps only bring marginal BLEU gains, while PrismDiff remains competitive with the best AR operating points over a wide latency range.

The key distinction between the two frontiers is the underlying trade-off mechanisms. 
To achieve faster inference, AR models must reduce decoder depth (model capacity), leading to a substantial drop in BLEU. 
In contrast, diffusion-based systems obtain faster operating points by reducing the number of refinement steps at inference time, without sacrificing model capacity. 
The left portion of the curve highlights this advantage: under strict low-latency budgets, PrismDiff gracefully maintains near-peak translation quality ($\sim$37 BLEU), whereas the AR frontier degrades sharply. 
Hence, this comparison does not claim that diffusion is superior to AR in every supervised setting; rather, it shows that diffusion systems provide a competitive performance frontier and a flexible control knob.

\subsection{Source-Side Zero-Shot Transfer}
The architectural advantages become even more pronounced under source-side zero-shot transfer. As shown in Figure~\ref{fig:frontier}, when evaluating the English-trained systems on unseen Spanish sources (Es-X), the PrismDiff frontier consistently outperforms the AR baselines across all points. 
Because both models utilize the identical source-side semantic representation, this performance gap cannot be attributed to differences in input features. 
Instead, it suggests that iterative refinement over a shared semantic anchor is more robust to source-side domain shift than traditional left-to-right decoding. 

We speculate that this is due to the cumulative effect of generation errors. In AR models, out-of-distribution inputs may trigger early decoding errors, which can cascade due to autoregression. 
In contrast, PrismDiff's bidirectional context and iterative refinement mitigate such error accumulation, resulting in a more stable generation process. 

This robustness is important for real-world deployment, where models trained on high-resource source languages encounter input from unseen sources. 
While PrismDiff does not completely bridge the supervised-to-zero-shot gap, 
it shows a stronger zero-shot frontier, suggesting that iterative refinement can more effectively exploit the semantic meaning regardless of the source language.

\subsection{Ablation Studies}
\paragraph{Effect of Semantic Anchor.} 
Figure~\ref{fig:frontier} and Table \ref{tab:multi30k} isolate the effect of the anchor from increasing $T$. 
A Diff-SrcTokens model trained with $T=10$ peaks at 33.84 BLEU; increasing to $T=100$ improves this to 35.07.
Under the same $T=100$ setting, PrismDiff consistently outperforms Diff-SrcTokens across matched latency budgets in supervised scenarios, confirming that a larger $T$ improves the sampling frontier but cannot fully explain PrismDiff's gains. 
The most important divergence emerges in zero-shot scenarios: without the anchor, Diff-SrcTokens degrades severely (BLEU$<1$ on Es-X), whereas PrismDiff maintains strong performance. 
Thus, a larger $T$ and the semantic anchor play complementary roles: the former improves latency-quality trade-offs, whereas the latter provides the necessary cross-lingual mapping for zero-shot transfer.

Interestingly, PrismDiff performs better with exclusive attention than full attention, whereas Diff-SrcTokens shows the opposite trend. 
This suggests that without the anchor, cross-stream attention serves as a necessary channel for implicit cross-lingual alignment; once the anchor fulfills this alignment, cross-stream attention becomes redundant and may introduce interference between language-specific surface forms. 
(See Table \ref{appendix_tab:multi30k_full})

\paragraph{Effect of Joint Multi-Stream Optimization.} 
While Diff-Indep uses the same diffusion mechanism and semantic anchor under matched $T=100$ setting, it  generates each target language independently. 
This ablation therefore clarifies whether PrismDiff's gains come solely from the anchored diffusion process. 
As shown in Figure~\ref{fig:frontier} and Table \ref{tab:multi30k}, PrismDiff consistently outperforms Diff-Indep in both supervised and zero-shot settings: at $t=30$, it maintains a $\sim$1.5 and $\sim$1.4 BLEU advantage over Diff-Indep on supervised and zero-shot setting, respectively. 
This suggests that jointly optimizing all streams provides not only deployment convenience for one-to-many translation, but also implicit cross-lingual regularization beyond what the anchor alone achieves. 

\paragraph{Effect of Reset Positional Encoding.} 
We also ablate the Reset Positional Encoding by replacing it with standard absolute positional encoding. 
As shown in Figure~\ref{fig:frontier}, this leads to a substantial performance drop (23.8 vs. 33.5 BLEU at $t=10$), indicating the necessity for the model to correctly identify token positions within each stream.

\begin{figure}[t]
\centering
  \includegraphics[width=1.0\columnwidth]{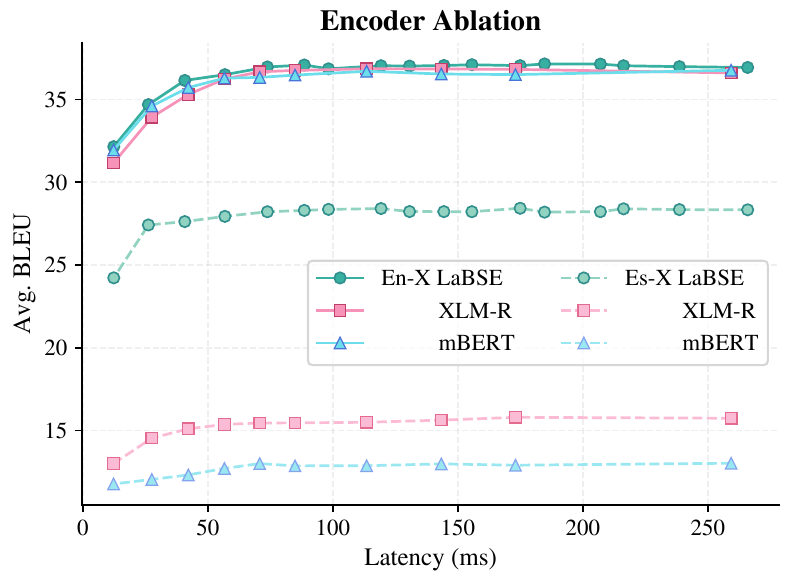}
\vspace{-6mm}
  \caption{\textbf{Encoder ablation} for PrismDiff ($T=100$) on Multi30K. Sold curves denote supervised En-X frontiers, and dashed curves denote zero-shot Es-X frontiers. 
  }
  \label{fig:encoder_ablation}
\vspace{-2mm}
\end{figure}

\subsection{Encoder Choice and Anchor Quality}
We further investigate whether the zero-shot transfer capability depends on the specific choice of multilingual encoder by comparing LaBSE against mBERT \citep{devlin-etal-2019-bert} and XLM-R \citep{conneau2020unsupervised}.

Figure~\ref{fig:encoder_ablation} demonstrates that the effectiveness of the semantic anchor relies on the specific properties of the multilingual encoder. 
When integrating LaBSE, mBERT, and XLM-R into the same PrismDiff architecture, all three encoders yield comparable performance on supervised En-X tasks.
However, their zero-shot Es-X frontiers diverge drastically. 
While LaBSE maintains robust zero-shot performance of near 28 BLEU, both XLM-R and mBERT collapse to significantly lower performance levels (near 13 and 10 BLEU, respectively). 

Importantly, we do not claim LaBSE is universally the best encoder choice across all decoding architectures. 
Experiments on AR models reveal that mBERT can achieve higher supervised quality ($\sim$39 BLEU) under traditional AR decoding, despite still failing catastrophically in Es-X zero-shot transfer ($\sim$8.5 BLEU). 
This suggests that the optimal encoder choice is dependent on both the decoding mechanism and the specific task. 
For PrismDiff, specifically, LaBSE is more effective because its language-agnostic sentence-level representations perfectly meet the need of a shared anchor to guide the iterative diffusion process.

\subsection{Deployment Flexibility and Scaling}
Our experiments further validate PrismDiff's architectural flexibility for real-world deployment. 
First, the framework is inherently invariant to the generation order of target languages. For example, generating En-\{De, Fr\} versus En-\{Fr, De\} yields identical performance.
Second, dynamically omitting target streams at inference time impacts translation quality by less than 1\%, proving that the system allows for flexible configuration without retraining. 
A core motivation of our framework is efficient one-to-many generation. We analyze the latency and memory scaling behavior as the number of target languages ($K$) increases (see \ref{appendix:scaling_analysis}). Traditional AR systems suffer from strict linear latency scaling, as generating $K$ languages sequentially multiplies the temporal bottleneck. In contrast, although the canvas expands with more target streams, PrismDiff exhibits a sublinear latency growth thanks to the parallel diffusion refinement. 


\section{Conclusion}
In this work, we explore the design space of parallel multilingual machine translation.
To illustrate this, we propose PrismDiff, a discrete diffusion framework for parallel one-to-many translation. 
By guiding the generative process with a language-agnostic semantic anchor, PrismDiff enables parallel one-to-many translation with high semantic consistency, controllable quality-latency trade-offs, and robust source-side zero-shot transfer. 
These results suggest that parallel multi-stream diffusion with a semantic anchor is a viable and flexible approach that offers a different quality-latency trade-off compared to AR systems. 
In supervised settings, PrismDiff with accelerated sampling achieves comparable translation quality to strong AR baselines while delivering a 2$\times$ speedup. More notably, PrismDiff establishes a stronger zero-shot frontier: the semantic anchor design enables the framework to maintain approximately 75\% of its supervised performance on unseen source languages, outperforming its AR counterparts in zero-shot BLEU.

Looking forward, our framework holds the potential to generalize to other one-to-many or many-to-many generation tasks beyond translation. Exploring more robust anchor alignment strategies could further improve performance.


\section*{Limitations}
Despite its advantages in latency and flexibility, PrismDiff has several limitations. 

First, as a diffusion-based NAR model, it does not uniformly outperform strong AR baselines on supervised translation quality. Its advantage lies in a controllable frontier: fewer refinement steps reduce latency, while additional steps enhance quality. Diffusion decoding also remains slower than single-step NAR models, necessitating further acceleration strategies. 

Second, our current approach relies on a pretrained language-agnostic sentence encoder (LaBSE) to extract the semantic anchor. Although this anchor enables strong zero-shot transfer in PrismDiff, encoder ablations show that not every multilingual encoder provides a sufficiently aligned semantic space. A poorly represented language, domain, or a truly distant low-resource language in the pretrained encoder may greatly degrade transfer performance. Moreover, the best source representation can be decoder-dependent; different multilingual encoders may benefit different decoder architectures. Thus, our conclusion about LaBSE should be interpreted within PrismDiff rather than as a universal claim for all translation decoders. 

Third, for experiment simplicity, we employ several vanilla implementation strategies of MDMs, such as simple noise schedules and fixed language block lengths. Although our settings ($l_\mathrm{multi30k}=40$, $l_\mathrm{europarl}=80$) can cover most examples, there still exist longer sentences that will overflow, which considerably contributes to the poor performance on Europarl. Better performance may be achieved if these strategies and hyperparameters are more carefully tuned. 

While exclusive attention already produces strong results through joint parameter optimization, exploring lightweight cross-stream interaction mechanisms, such as gated cross-attention between target streams, could be a promising direction but may also introduce more computation and inter-stream interference. 

Additionally, our efficiency evaluation focuses on a standard single-GPU, single-instance, batch-size=1 latency, which closely reflects serving settings with extremely high latency requirements where synchronous batch is impractical. Optimizing batch throughput remains a direction to be explored in the future.




\bibliography{paper}

\appendix

\section{Appendix}
\label{sec:appendix}



\subsection{Logarithm-Uniform Masking Schedule}
\label{appendix:log_schedule}
Let $L$ denote the maximum sequence length. The masking ratio $m$ is bounded within $[1/L, 1]$, where $1/L$ corresponds to the single-token mask limit. 
At each refinement step, tokens are remasked based on predicted confidence scores: the model retains the top $(1-m)\times L$ tokens with highest predicted probability and remasks the remainder, where $m$ is determined by the schedule. 
We define a log-uniform distribution over $m$ by sampling:
    $$\log m \sim \mathcal{U}\left(\log \frac{1}{L}, 0\right)$$
This formulation assigns equal probability mass to equal intervals on the logarithmic scale.
To adapt this continuous prior for fixed-step training, we discretize the schedule over the total training steps $N$. 
In practice, the log-uniform schedule is implemented deterministically by mapping training steps to masking ratios. 
For example, if using $L=40$ with $K=4$ discrete levels, the corresponding time step sequence $t_i$ is given by: 
    $$t_i= \exp \left( \log(\frac{1}{L})\times i \right) \times N$$
, which are: 
$$t_0 = N, \ t_1 = 0.4N, \ t_2 = 0.16N, \ t_3 = 0.063N$$
This ensures that the training process effectively covers the full range of masking ratios with the desired log-uniform distribution. 
We found this masking schedule consistently outperforms other simple strategies in our task.

\subsection{Scaling Analysis: Latency and Memory}
\label{appendix:scaling_analysis}

\begin{figure}[ht]
\centering
  \includegraphics[width=1.0\columnwidth]{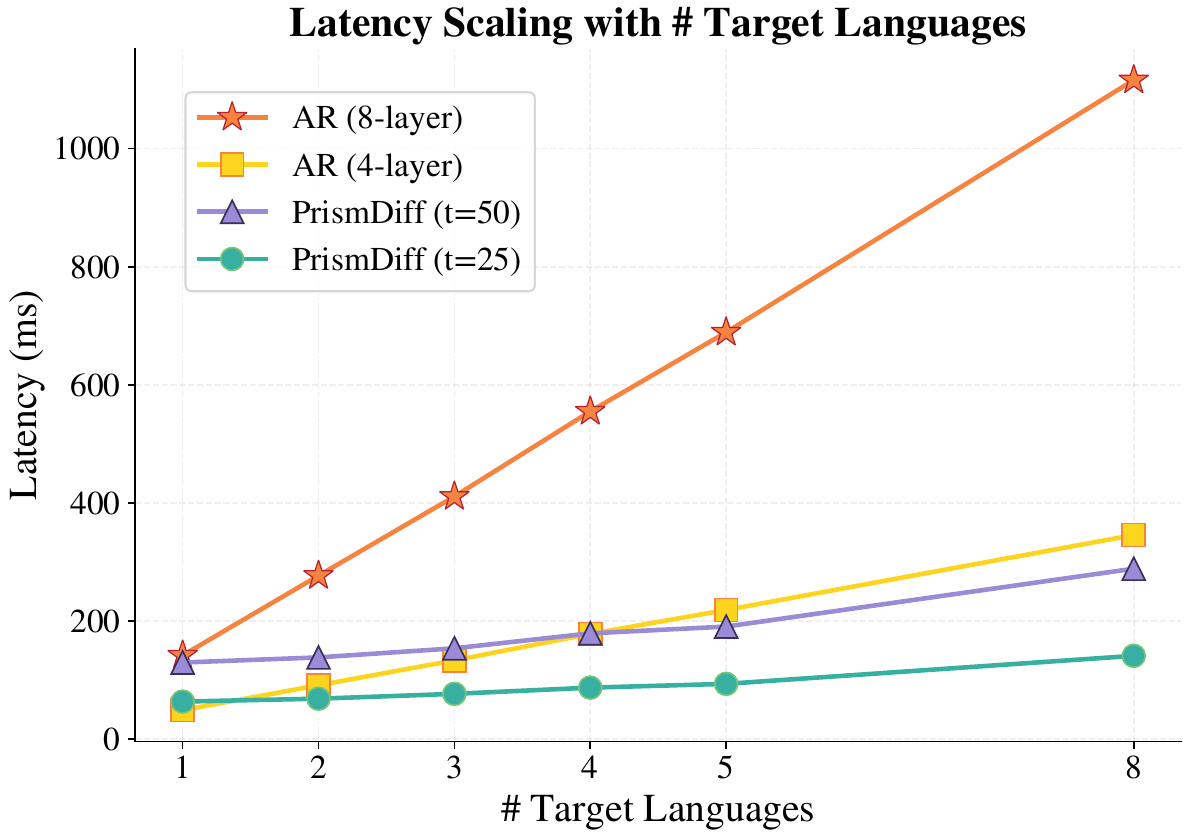}
  \caption{\textbf{Latency--language scaling graph.} The PrismDiff models are trained with $T=100$, and accelerated sampling is performed during the inference stage using $t=50$ or $t=25$. 
  }
  \label{appendix:fig:scale}
\end{figure}

A detailed latency--language scaling analysis is provided in Figure \ref{appendix:fig:scale}, comparing decoder-only AR (8-layer and 4-layer) against PrismDiff ($T=100$, $t=50$, and $t=25$). 
PrismDiff exhibits sublinear latency scaling with respect to the number of target streams $K$.
With exclusive attention, each refinement step requires one forward pass over the full canvas of size $L$, where each stream attends only to itself and the anchor with cost $O(L^2)$. 
The per-step cost is therefore $O(K \cdot L^2)$, and the total cost over $t$ diffusion refinement steps is $O(t \cdot K \cdot L^2)$.
Generating $K$ languages in a single forward pass therefore is $K\times$ the cost of a single-language diffusion pass, but $K\times$ cheaper than running $K$ separate passes. 

In contrast, AR systems decoding for $K$ languages requires $K$ sequential passes, each with $O(L^2)$ attention cost over $L$ autoregressive steps, resulting in $O(K\cdot L^3)$ in total. 
PrismDiff's total cost of $O(t \cdot K \cdot L^2)$ is therefore favorable when $t \ll L$. 
We note that AR systems could reduce wall-clock latency for $K$ targets by running $K$ model replicas simultaneously. However, this would multiply the memory consumption by $K$ whereas PrismDiff generates all $K$ targets with a single model instance.

\begin{table}[ht]
\centering
\setlength{\tabcolsep}{5pt}
\resizebox{1.0\columnwidth}{!}{%
    \begin{tabular}{lcccccc}
    \toprule
    \textbf{Block length} & \multicolumn{6}{c}{\textbf{Number of Target Languages ($K$)}} \\
    \cmidrule(lr){2-7}
    & 1 & 2 & 3 & 4 & 5 & 8 \\
    \midrule
    $l=40$  & 2.66 & 2.70 & 2.73 & 2.77 & 2.81 & 2.91 \\
    $l=100$ & 2.72 & 2.81 & 2.90 & 2.99 & 3.08 & 3.35 \\
    $l=300$  & 2.90 & 3.17 & 3.44 & 3.71 & 3.98 & 4.80 \\
    \bottomrule
    \end{tabular}
}
\caption{\textbf{Peak GPU memory footprint} (in GB) of PrismDiff during parallel multi-stream inference. The memory scales gracefully even under extreme stress tests (e.g., $K=8$ targets with $l=300$ tokens).} 
\label{appendix_tab:memory}
\end{table}

Table~\ref{appendix_tab:memory} reports peak GPU memory footprint (GB) of PrismDiff under different block lengths ($l$) and number of target streams ($K$). 
With $l=40$ block length, memory increases by less than 10\% as $K$ scales from 1 to 8 (2.66 GB to 2.91 GB), confirming that the canvas-based multi-stream design does not incur significant memory overhead in practice. 
The memory growth becomes more pronounced at larger block lengths (e.g., $l=300$), which is expected as the total canvas size scales as $L_\mathrm{src}+K\times L_\mathrm{tgt}$. 
The memory scales gracefully even under extreme stress tests (e.g., $K=8$ with $L=300$), confirming that while the hyper-sequence design inherently costs more memory for long texts, it easily fits in the capacity of standard GPUs without requiring complex multi-GPU model parallelism. 
Note that AR memory does not scale with $K$ since each language is decoded sequentially; for reference, an 8-layer decoder-only AR model on Multi30K (max length $<40$) has a peak memory of approximately 2.69 GB, comparable to PrismDiff at $K=1$ (2.66 GB). 
The key difference lies not in memory but in latency: the cost of AR models is reflected in a linear latency growth, as shown in Figure~\ref{appendix:fig:scale}.

\subsection{Full Results}
\label{appendix:results}
Full experimental results on Multi30K are reported in Tables \ref{appendix_tab:multi30k_full} -- \ref{appendix_tab:multi30k_compare_encoders}. The results include comparing different implementation setups and architectures (Table \ref{appendix_tab:multi30k_full}), using different sampling budgets (Table \ref{appendix_tab:multi30k_steps}), decoder-only AR baselines with various numbers of decoder layers (Table \ref{appendix_tab:multi30k_AR_layers}), and applying different multilingual encoders for extracting source-side representation (anchor) as the input of PrismDiff (Table \ref{appendix_tab:multi30k_compare_encoders}).

\subsection{The Role of Cross-Lingual Alignment in Attention}
The performance gap of Diff-Indep using full attention versus using exclusive attention suggests that full cross-stream attention plays different roles depending on the conditioning signal. 
When conditioned directly on source tokens (Diff-SrcTokens), the model lacks a centralized, language-agnostic representation. 
Without a semantic anchor, full attention can partially compensate for weak source conditioning: each target stream essentially treats the noisy states of other streams as auxiliary context, forming a pseudo multi-source prediction problem to extract semantic clues. 
However, when PrismDiff is conditioned on a cross-lingually aligned semantic anchor, direct target-stream attention can introduce surface-form interference across languages. 
Therefore, exclusive stream attention encourages each language to perform its own denoising process, while semantic coordination is pre-solved by the shared anchor.

\subsection{Inference Examples}
One example of the accelerated inference process of PrismDiff trained on Multi30K with a total diffusion steps of $T=100$ is shown in Table \ref{tab:example1}, and Table \ref{tab:example2} contains another example trained on Europarl with $T=40$. 
All target languages are refined simultaneously and generated together at each step. 
The canvas size and the refinement steps of each target language is truncated in the examples for better readability. A larger step $t$ means an earlier stage in the inference time, so more mask tokens ($\texttt{[M]}$) will be presented in the example. 
The translation displayed in the last step ($t=0$) is CTC-collapsed $t=0$ prediction, which removes repetitive tokens (only if there exist).

\subsection{Zero-Shot Transfer Examples} 
We present representative examples illustrating typical failure modes of AR systems under zero-shot transfer in Table \ref{appendix_tab:zero_shot_examples}. 
We provide a qualitative analysis using zero-shot Spanish-to-French (Es $\rightarrow$ Fr) translation examples from Multi30K. The original task on which the model is trained on is En-\{De, Fr\}. 
In Example 1, the AR model suffers from error cascading and hallucination. In Example 2 and Example 3, the AR baselines exhibits structural collapse, generating non-existing words.

\subsection{Implementation of TET} 

Our implemented TET model on Europarl contains 4 leaf nodes (corresponding to the target languages), each has a depth of 6. 
We list the detailed paths of all languages in TET in Table \ref{appendix_tab:tet_path}. 
In all, the TET model contains 16 Transformer encoder layers, resulting in a total parameter count of 556M. 

TET's extremely low latency is attributed to its single-pass CTC decoding. Unlike sequential autoregressive decoding, or iterative diffusion refinements, TET generates all target languages in a single forward pass. The higher parameter count does not proportionally increase latency due to the parallel tree structure. 

\begin{table}[ht]
\centering
\setlength{\tabcolsep}{3pt}
\resizebox{\columnwidth}{!}{%
    \begin{tabular}{c|l}
    \toprule
     \textbf{Language} & \multicolumn{1}{c}{\textbf{Path}} \\ 
    \midrule  
      Dutch & \textit{--Germanic--Western--Dutch$_1$--Dutch$_2$}  \\ 
      Danish & \textit{--Germanic--Northern--Danish$_1$--Danish$_2$}  \\ 
      French & \textit{--Romance--Western--French$_1$--French$_2$}  \\ 
      Romanian & \textit{--Romance--Eastern--Romanian$_1$--Romanian$_2$}  \\ 
    \bottomrule
    \end{tabular}
}
\caption{Paths of all languages in TET on Europarl. 
All paths begin with "\textit{common$_1$--common$_2$--$\cdots$}'', which is omitted in this table for simplicity. }
\label{appendix_tab:tet_path}
\end{table}

\subsection{Licenses and Usage of Artifacts} 
The Multi30K dataset is publicly available and licensed under the CC BY-SA 4.0. Europarl dataset is publicly available as open data and widely permitted for research purposes. 
We utilize the pre-trained LaBSE (Language-Agnostic BERT Sentence Embeddings) and mBERT (multilingual BERT for model design and evaluation, both of which are officially released by Google and distributed under the Apache License 2.0. We also employ XLM-R (XLM-RoBERTa), developed by Meta, which is released under the MIT License. We also use the NLLB (No Language Left Behind, e.g., NLLB-200) model, which is distributed by Meta under the CC-BY-NC 4.0, strictly limiting its use to non-commercial research purposes. 
We confirm that our use of the datasets is consistent with their intended use for machine translation research. Similarly, all models are employed for feature extraction and data synthesis, respectively, aligning with their original research purposes.

\begin{table*}[ht]
\centering
\setlength{\tabcolsep}{5pt}
\resizebox{1.0\textwidth}{!}{%
  \begin{tabular}{cllcc||cc|cc||cc|r}
    \toprule
        
&&    &&    &  \multicolumn{2}{c|}{\textbf{BLEU}} & \multicolumn{2}{c||}{\textbf{COMET}}  & \multicolumn{2}{c|}{\textbf{Avg.}} & \multicolumn{1}{c}{\textbf{Latency}}\\
     \textit{Src.} 
&\textbf{Model} &  Architecture &Attn.  &  $T$&   De & Fr & De & Fr & \textbf{BLEU} & \textbf{COMET} &\multicolumn{1}{c}{(ms)}\\
    \midrule
     \textit{En} 
&Diff-Indep  &     &&  10&   28.47& 39.51& 0.62& 0.73& 33.99& 0.68& 26.9\\
 & & & & 100& 31.07& 40.74& 0.66& 0.76& 35.91& 0.71&254.5\\
  
\cmidrule{2-12}
&Diff-SrcTokens   &  &full&  10& 30.35& 37.32& 0.61& 0.70& 33.84& 0.66&25.6\\
 & & - CTC& full& 10& 29.05& 36.31& 0.58& 0.67& 32.68& 0.63&24.5\\
 & & & full& 100& 32.12& 40.76& 0.65& 0.74& 36.44& 0.70&234.4\\
 & & & excl.& 100& 30.95& 38.64& 0.65& 0.74& 34.80& 0.695&247.4\\

\cmidrule{2-12}
&AR&     enc-dec &&   &   29.90& 35.83& 0.67& 0.73& 32.87& 0.70& 88.9\\
 & & enc-dec + Z& & & 20.85& 22.65& 0.61& 0.66& 21.75& 0.64&90.1\\

&&  dec-only + Z&&  & 32.82& 41.67& 0.68& 0.76& 37.25& 0.72&148.3\\
    \cmidrule{2-12}
&PrismDiff&  &full &   10&   27.91& 37.40& 0.61& 0.71& 32.66& 0.66& 27.8\\
 & & - CTC& full& 10& 27.16& 36.62& 0.58& 0.68& 31.89& 0.63&32.5\\
     
 & & & excl.& 10& 28.48& 38.47& 0.62& 0.72& 33.48& 0.67&26.5\\
 & & - CTC& excl.& 10& 27.29& 37.09& 0.59& 0.70& 32.19& 0.65&26.7\\
 & & Abs. PE& excl.& 10& 19.56& 28.04& 0.53& 0.62& 23.80& 0.58&26.2\\
 & & & full& 100& 31.92& 40.24& 0.67& 0.75& 36.08& 0.71&259.0\\
  
&&  & excl.&  100& 33.16& 41.10& 0.68& 0.76& 37.13& 0.72&278.3\\
    \midrule \midrule
     \textit{Es} 
&Diff-SrcTokens   &    &full&  10&   0.28& 0.16& 0.28& 0.29& 0.22& 0.29&  21.9\\
 & & & full& 100& 0.14& 0.31& 0.29& 0.30& 0.23& 0.30&234.0\\
 & & & excl.& 100& 0.30& 0.31& 0.26& 0.32& 0.31& 0.29&238.6\\
  
\cmidrule{2-12}
&AR&  dec-only + Z&&  & 20.48& 29.95& 0.60& 0.71& 25.22& 0.66&154.5\\
    \cmidrule{2-12}
&PrismDiff&  &full&  10& 19.43& 30.43& 0.56& 0.68& 24.93& 0.62&29.5\\
  
 & & & excl.& 10& 20.33& 31.51& 0.57& 0.69& 25.92& 0.63&27.2\\
 & & Abs. PE& excl.& 10& 14.75& 21.53& 0.50& 0.59& 18.14& 0.55&25.8\\
 & & & full& 100& 22.37& 33.13& 0.62& 0.73& 27.75& 0.67&25.6\\
     &&  &excl.&   100&   23.25& 33.17& 0.62& 0.73& 28.21& 0.68&  284.0\\
    \bottomrule
  \end{tabular}
}
\caption{Results on Multi30K. 
In architecture ablations, ``enc-dec'' is encoder-decoder and ``dec-only'' denotes decoder-only;  ``+Z'' means using semantic anchor Z as input, otherwise the AR model directly uses source tokens as input; ``-CTC'' means without CTC decoding; ``Abs. PE'' is using absolute positional encoding instead of reset positional encoding. 
``Attn.'' represents the attention mode used by PrismDiff, in which ``full'' means using full attention across target language, and ``excl.'' means using exclusive attention for each target language. 
$T$ is the total diffusion steps the model is trained on, and all diffusion systems report the results using $T$ full sampling steps with uniform noise scheduler. 
``Src.'' lists the source language input to the model, in which Spanish (Es) is a zero-shot source language that the model is never trained on.}
\label{appendix_tab:multi30k_full}
\end{table*}

\begin{table*}[ht]
\centering
\setlength{\tabcolsep}{5pt}
\resizebox{0.75\textwidth}{!}{%
\begin{tabular}{cc||cc|cc||cc|r}
    \toprule
      
& & \multicolumn{2}{c|}{\textbf{BLEU}} & \multicolumn{2}{c||}{\textbf{COMET}} & \multicolumn{2}{c|}{\textbf{Avg.}} & \multicolumn{1}{c}{\textbf{Latency}}
\\
     \textit{Src.}
&Sampl. Steps ($t$)& De & Fr & De & Fr & \textbf{BLEU} & \textbf{COMET} & \multicolumn{1}{c}{(ms)} \\
    \midrule 
     \textit{En} 
&100 & 33.16 & 41.10 & 0.68 & 0.76 & 37.13& 0.72& 278.3 \\
     
&50& 32.90 & 41.22 & 0.68 & 0.76 & 37.06& 0.72& 147.2 \\
     
&30 & 33.20 & 40.98 & 0.67 & 0.76 & 37.09& 0.72& 83.2 \\
     
&25 & 32.99 & 40.95 & 0.67 & 0.76 & 36.97& 0.72& 68.7 \\
     
&15& 32.30 & 40.01 & 0.67 & 0.75 & 36.16& 0.71& 40.9 \\
    \midrule \midrule 
     \textit{Es} 
&100 & 23.25 & 33.17 & 0.62 & 0.73 & 28.21& 0.68& 284.0\\
     
&50& 23.44 & 33.01 & 0.62 & 0.73 & 28.23& 0.67& 141.7 \\
     
&30 & 23.69 & 32.91 & 0.62 & 0.72 & 28.30& 0.67& 83.1 \\
     
&25 & 23.43 & 33.01 & 0.62 & 0.72 & 28.22& 0.67& 69.3 \\
 &15& 22.91 & 32.34 & 0.61 & 0.71 & 27.63& 0.66& 40.7 \\
    \bottomrule
\end{tabular}
}
\caption{Sampling steps vs. performance on Multi30K. The PrismDiff model is trained with a total diffusion steps of $T=100$, and is evaluated under different accelerated sampling steps ($t$) using logarithm-uniform noise scheduler.}
\label{appendix_tab:multi30k_steps}
\end{table*}

\begin{table*}[ht]
\centering
\setlength{\tabcolsep}{5pt}
\resizebox{0.75\textwidth}{!}{%
\begin{tabular}{lc||cc|cc||cc|r}
    \toprule
      
& & \multicolumn{2}{c|}{\textbf{BLEU}} & \multicolumn{2}{c||}{\textbf{COMET}} & \multicolumn{2}{c|}{\textbf{Avg.}} &  \multicolumn{1}{c}{\textbf{Latency}}
\\
     \textit{Src.}
&Num. Layers & De & Fr & De & Fr & \textbf{BLEU} & \textbf{COMET} & \multicolumn{1}{c}{(ms)} \\
    \midrule 
     \textit{En} 
&8& 32.82& 41.67& 0.68& 0.76& 37.25& 0.720& 148.3\\
&6& 32.36& 40.45& 0.70& 0.77& 36.41& 0.735& 120.9\\ 
&4& 31.97& 40.08& 0.68& 0.76& 36.03& 0.720& 92.5\\  
&3& 30.79& 40.16& 0.67& 0.75& 35.48& 0.710& 76.0\\
&2& 30.96& 38.77& 0.66& 0.75& 34.87& 0.705& 58.6\\
    \midrule \midrule 
     \textit{Es} 
&8& 20.48& 29.95& 0.60& 0.71& 25.22& 0.655& 154.5\\
&6& 20.98& 30.32& 0.63& 0.73& 25.65& 0.678& 121.4\\  
&4& 21.67& 30.39& 0.62& 0.72& 26.03& 0.670& 92.5\\
&3& 19.73& 29.35& 0.60& 0.71& 24.54& 0.655& 77.6\\
&2& 19.30& 28.82& 0.59& 0.70& 24.06& 0.645& 61.0\\
    \bottomrule
\end{tabular}
}
\caption{Number of AR decoder layers vs. performance on Multi30K. }
\label{appendix_tab:multi30k_AR_layers}
\end{table*}

\begin{table*}[ht]
\centering
\setlength{\tabcolsep}{5pt}
\resizebox{0.85\textwidth}{!}{%
\begin{tabular}{clc||cc|cc||cc|r}
    \toprule
      
&&   & \multicolumn{2}{c|}{\textbf{BLEU}} & \multicolumn{2}{c||}{\textbf{COMET}} & \multicolumn{2}{c|}{\textbf{Avg.}} &  \multicolumn{1}{c}{\textbf{Latency}}
\\
     \textit{Src.}
&Encoder&  Sampl. Steps ($t$)& De & Fr & De & Fr & \textbf{BLEU} & \textbf{COMET} & \multicolumn{1}{c}{(ms)} \\
    \midrule 
     \textit{En}
&LaBSE&  100& 33.16& 41.10& 0.68& 0.76& 37.13& 0.72& 376.8\\
  
&& 30& 33.20& 40.98& 0.67& 0.76& 37.09& 0.72&83.2\\
  
&& 25& 32.99& 40.95& 0.67& 0.76& 36.97& 0.72&68.7\\
  
&& 15& 32.30& 40.01& 0.67& 0.75& 36.16& 0.71&40.9\\
    \cmidrule{2-10}
 
&mBERT&   100& 33.44& 40.55& 0.67& 0.75& 37.00& 0.71& 281.2\\
  
&& 30& 33.00& 39.96& 0.67& 0.75& 36.48& 0.71&82.9\\
  
&& 25& 32.64& 40.04& 0.67& 0.75& 36.34& 0.71&69.4\\
  
&& 15& 32.34& 39.11& 0.66& 0.74& 35.73& 0.7.0&40.1\\
    \cmidrule{2-10}
 
&XLM-R&   100& 33.45& 40.20& 0.68& 0.76& 36.83& 0.72& 286.0\\
  
&& 30& 33.43& 40.09& 0.67& 0.76& 36.76& 0.72&86.8\\
  
&& 25& 33.52& 39.84& 0.67& 0.76& 36.68& 0.72&72.8\\
  
&& 15& 31.62& 39.00& 0.66& 0.75& 35.31& 0.71&44.6\\
    \midrule \midrule 
     \textit{Es} 
&LaBSE&  100& 23.25& 33.17& 0.62& 0.73& 28.21& 0.68& 284.0\\
  
&& 30& 23.69& 32.91& 0.62& 0.72& 28.30& 0.67&83.1\\
  
&& 25& 23.43& 33.01& 0.62& 0.72& 28.22& 0.67&69.3\\
  
&& 15& 22.91& 32.34& 0.61& 0.71& 27.63& 0.66&40.7\\
    \cmidrule{2-10}
&mBERT&   100& 10.10& 16.06& 0.47& 0.54& 13.08& 0.51& 283.5\\
  
&& 30& 9.79& 15.93& 0.47& 0.53& 12.86& 0.50&82.8\\
  
&& 25& 10.13& 15.87& 0.47& 0.53& 13.00& 0.50&69.2\\
  
&& 15& 9.63& 15.00& 0.46& 0.52& 12.32& 0.49&40.0\\
    \cmidrule{2-10}
  
&XLM-R& 100& 13.68& 18.18& 0.51& 0.59& 15.93& 0.55&289.5\\
  
&& 30& 13.25& 17.67& 0.51& 0.59& 15.46& 0.55&87.1\\
  
&& 25& 13.04& 17.87& 0.51& 0.59& 15.46& 0.55&71.7\\
     &&   15& 12.74& 17.46& 0.51& 0.58& 15.10& 0.55& 44.2\\
    \bottomrule
\end{tabular}
}
\caption{Performance comparison of using different multilingual encoders to extract the source-side representation (anchor) for PrismDiff on Multi30K. All PrismDiff models are trained with total diffusion steps of $T=100$.}
\label{appendix_tab:multi30k_compare_encoders}
\end{table*}

\begin{table*}[ht]
\centering
\begin{tblr}{
  colspec = {Q[l] X[l] X[l] }, 
  width = \textwidth,
  hlines = {0.3pt},
  hline{1,Z} = {1.2pt},
  vlines = {0.3pt},
  vline{1,Z} = {0pt},
  rows = {t},
  row{1} = {font=\normalsize},
  row{2-Z} = {font=\small},
}
    \textbf{Step} & \textbf{German} & \textbf{French} \\
    t=54.1 & Ein [M] [M] [M] [M] [M] [M] [M] [M] [M] [M] [M] & Un homme [M] [M] [M] [M] [M] [M] [M] [M] [M] [M] [M] \\
    t=18.4 & Ein Mann schläft in einem grünen [M] auf [M] [M]fa [M] & Un homme dormant dans une [M] [M] un can [M]é [M] [M] \\
    t=5.4 & Ein Mann schläft in einem grünen [M] auf einem Sofa. & Un homme dormant dans une pièce sur un canapé [M]. \\
    t=0 & Ein Mann schläft in einem grünen Zimmer auf einem Sofa. & Un homme dormant dans une pièce sur un canapé vert. \\
    \hline[\heavyrulewidth]
    \textbf{Groundtruth} & Ein Mann schläft in einem grünen Raum auf einer Couch. & Un homme dormant dans une chambre verte sur un canapé.  \\
\end{tblr}
\caption{An example of accelerated (with 25 refinement steps) translation process of PrismDiff ($T=100$) trained on Multi30K. The canvas size and the number of the steps are reduced in this example for better readability. The ``t=0'' is the result after CTC collapse. The source English sentence is: ``\textit{A man sleeping in a green room on a couch.}''}
\label{tab:example1}
\end{table*}

\begin{table*}[ht]
\centering
\begin{tblr}{
  colspec = {Q[l] X[l] X[l] X[l]}, 
  width = \textwidth,
  hlines = {0.3pt},
  hline{1,Z} = {1.2pt},
  vlines = {0.3pt},
  vline{1,Z} = {0pt},
  rows = {t},
  row{1} = {font=\normalsize},
  row{2-Z} = {font=\small},
}
    \textbf{Step} & \textbf{French} & \textbf{Romanian} & \textbf{Danish} \\
    t=30
        & Le Parlement européen [M] [M] [M] [M] [M] [M] [M] [M] [M] [M] [M] [M] [M] [M] 
        & Parlamentul European a votat [M] [M] [M] [M] [M] [M] [M] [M] [M] [M] [M] [M] [M] [M]
        & [M] [M] [M] [M] [M] [M] [M] [M] [M] [M] [M] [M] [M] [M] [M] [M] [M] [M] [M] [M] 
        \\
    t=20
        & Le Parlement européen a voté [M] [M] résolution [M] [M] [M] [M] [M] [M] [M] [M] 
        & Parlamentul European a votat o rezolu [M] [M] [M] [M] [M] [M] [M] [M] [M] [M] [M] [M]
        & Parlamentet har stemt for [M] beslutning om [M] [M] [M] [M] [M] [M] [M] [M] [M] [M] [M]
        \\
    t=10
        & Le Parlement européen a voté pour [M] résolution [M] l'avenir du Fonds social [M] [M]
        & Parlamentul European a votat [M] rezolu [M] referitoare la viitorul Fondului [M] [M] [M]
        & Parlamentet har stemt for en beslutning om Den Europæiske Social [M] [M] [M] [M] [M]
        \\
    t=0 
        & Le Parlement européen a voté pour une résolution sur l'avenir du Fonds social européen.
        & Parlamentul European a votat o rezoluție referitoare la viitorul Fondului social european.
        & Parlamentet har stemt for en beslutning om Den Europæiske Socialfonds fremtid.
        \\
    \hline[\heavyrulewidth]
    \textbf{Groundtruth} 
    & Le Parlement européen a voté une résolution sur l'avenir du Fonds social européen.
    & Parlamentul European a votat pentru o rezoluție referitoare la viitorul Fondului social european.
    & Parlamentet har stemt for beslutningen om Den Europæiske Socialfonds fremtid.
    \\
\end{tblr}
\caption{An example of translation process of PrismDiff ($T=40$) trained on Europarl. Here, only 3 target languages are displayed for simplicity. The canvas size and the number of the steps are reduced in this example for better readability. The ``t=0'' is the result after CTC collapse. The source English sentence is: ``\textit{The European Parliament has voted for a resolution on the future of the European Social Fund.}''}
\label{tab:example2}
\end{table*}

\begin{table*}[ht]
\centering
\setlength{\tabcolsep}{5pt}
\resizebox{0.85\textwidth}{!}{%
    \begin{tabular}{l|l}
    \toprule 
    
     \textbf{Example 1}& \multicolumn{1}{c}{ } \\ 
    \midrule  
      Source (En)& \textit{A child is splashing in the water}\\ 
    \midrule
      Source (Es)& \textit{Un niño está salpicando en el agua}\\
    \midrule
 Groundtruth (Fr)&"Un enfant se plonge dans l'eau"\\ 
      AR& \textit{Un enfant est en train de donner de l'eau en train de shooter dans l'eau}\\ 
      PrismDiff& \textit{Un enfant fait des éclaboussures dans l'eau}\\ 
      
    \midrule \midrule
    
    \textbf{Example 2}& \multicolumn{1}{c}{ } \\ 
    \midrule  
      Source (En)& \textit{This lady has heard a funny joke and laughing}\\ 
    \midrule
      Source (Es)& \textit{Esta señora ha oído un chiste divertido y riéndose}\\
    \midrule
 Groundtruth (Fr)&\textit{Cette dame a entendu une blague drôle et ri}\\ 
      AR& \textit{Cette femme a une ribe riée et rit riant}\\ 
      PrismDiff& \textit{Cette femme écoute une drôle de rire et rire}\\ 
          
    \midrule \midrule
    
    \textbf{Example 3}& \multicolumn{1}{c}{ } \\ 
    \midrule  
      Source (En)& \textit{A man holding up another with his back}\\ 
    \midrule
      Source (Es)& \textit{Un hombre sosteniendo a otro con la espalda}\\
    \midrule
 Groundtruth (Fr)&\textit{Un homme qui tient un autre avec le dos}\\ 
      AR& \textit{Un homme tient un autre ab qui avec l'arrière}\\ 
      PrismDiff& \textit{Un homme se tenant un autre avec le dos}\\ 
      
    \bottomrule
    \end{tabular}
}
\caption{Zero-shot examples on Multi30K. 
The input source is Spanish (Es), and target languages are German (De) and French (Fr). We only show French sentences for illustration. The English source sentences in this table are only for reference (not used as input). }
\label{appendix_tab:zero_shot_examples}
\end{table*}

\end{document}